\documentclass[10pt, conference, compsocconf]{IEEEtran}

\usepackage{cite}
\usepackage{amsmath,amssymb,amsfonts}
\usepackage{algorithmic}
\usepackage{graphicx}
\usepackage{textcomp}
\usepackage{pdfpages}
\usepackage{hyperref}
\usepackage{multirow}
\usepackage{subfigure}
\usepackage{epsfig}
\usepackage{gensymb}
\usepackage{tikz}
\usepackage{tabu}
\usepackage{booktabs,tabularx}
\usepackage{array}
\usepackage[nolist,nohyperlinks]{acronym}
\usepackage{cleveref}

\usetikzlibrary{shapes.geometric,shapes.arrows,decorations.pathmorphing}
\usetikzlibrary{matrix,chains,scopes,positioning,arrows,fit}

\crefname{figure}{figure}{figures}
\Crefname{figure}{Figure}{Figures}
\renewcommand{\figurename}{Figure}

\acrodef{AUC}{area under the blood-concentration versus time curve}
\acrodef{TDM}{therapeutic drug monitoring}
\acrodef{ANN}{artificial neural network}
\acrodef{ML}{machine learning}
\acrodef{XAI}{explainable artificial intelligence}
\acrodef{PK}{pharmacokinetic}
\acrodef{RMSE}{root mean squared error}
\acrodef{SHAP}{Shapley additive explanations}
\acrodef{TXT}{time after transplantation}
\acrodef{PE}{predictive error}
\acrodef{CV}{cross-validation}

\begin{document}
\title{Predicting tacrolimus exposure in kidney transplanted patients using machine learning}




 \newcommand{\simulamet}{SimulaMet, Norway}
 \newcommand{\oslomet}{Oslo Metropolitan University, Norway}
 \newcommand{\ntnu}{Norwegian University of Science and Technology, Norway}
 \newcommand{\riksen}{Rikshospitalet, Oslo University Hospital, Norway}
 \newcommand{\uio}{University of Oslo, Norway}
 \newcommand{\uit}{UiT The Arctic University of Norway}
 \author{
     \IEEEauthorblockN{
         Andrea M.\ Stor{\aa}s \IEEEauthorrefmark{1}\IEEEauthorrefmark{2},
         Anders {\AA}sberg\IEEEauthorrefmark{3}\IEEEauthorrefmark{4},
         P{\aa}l Halvorsen\IEEEauthorrefmark{1}\IEEEauthorrefmark{2},
         Michael A.\ Riegler\IEEEauthorrefmark{1}\IEEEauthorrefmark{9},
         Inga Str{\"u}mke\IEEEauthorrefmark{1}\IEEEauthorrefmark{5}
     }
     \IEEEauthorblockA{%
         \IEEEauthorrefmark{1}\simulamet \ \ 
         \IEEEauthorrefmark{2}\oslomet  \ \ 
         \IEEEauthorrefmark{5}\ntnu \\ 
         \IEEEauthorrefmark{3}\riksen  \ \
         \IEEEauthorrefmark{4}\uio  \ \ 
         \IEEEauthorrefmark{9}\uit  
     }
}

\tikzstyle{block} = [rectangle, draw, fill=blue!20, text width=5em, text centered, rounded corners, minimum height=4em, node distance=3cm]
\tikzstyle{line} = [draw, -latex']
\tikzstyle{image} = [inner sep=0pt, node distance=3cm]


\maketitle

\begin{abstract}
Tacrolimus is one of the cornerstone immunosuppressive drugs in most transplantation centers worldwide following solid organ transplantation. Therapeutic drug monitoring of tacrolimus is necessary in order to avoid rejection of the transplanted organ or severe side effects. However, finding the right dose for a given patient is challenging, even for experienced clinicians. Consequently, a tool that can accurately estimate the drug exposure for individual dose adaptions would be of high clinical value. In this work, we propose a new technique using machine learning to estimate the tacrolimus exposure in kidney transplant recipients. Our models achieve predictive errors that are at the same level as an established population pharmacokinetic model, but are faster to develop and require less knowledge about the pharmacokinetic properties of the drug.
\end{abstract}


\section{Introduction}
The preferred treatment of end-stage kidney disease is transplantation, with better long-term outcomes and patient quality of life compared to dialysis for the selected patients that are eligible for a transplantation~\cite{abecassis2008kidneyTx}. Individually adapted immunosuppressive therapy is a key factor for optimal long-term outcomes after kidney transplantation. Maintenance immunosuppressive therapy usually consists of two to three immunosuppressive drugs in combination, with individually tailored doses to obtain exposure within prespecified therapeutic windows. Tacrolimus, a calcineurin inhibitor, is the backbone of most standard protocols world-wide following kidney transplantation~\cite{brunet2019consensus}. Tacrolimus dosing is challenging even for experienced clinicians due to high individual variability in drug exposure. Standard \ac{TDM} is performed based on measuring trough concentrations, i.e., the concentration immediately prior to the subsequent dosing. However, recent consensus guidelines recommend a transfer to target systemic exposure measures for tacrolimus \ac{TDM}, i.e., the \ac{AUC} within a dose interval. Consequently, a tool that accurately predicts the systemic exposure of tacrolimus based on a reasonable number of blood samples for a given patient would be of high clinical value.    

The aim of this study is to develop a \ac{ML} model for predicting the systemic tacrolimus exposure in kidney transplanted patients utilizing a limited sampling strategy applicable in clinical practice. 
We aim that the model, which is trained using XGBoost~\cite{Chen2016xgboost}, is robust to varying blood sampling time schedules, increasing the model's usability for real-life settings. Our model is prospectively externally validated in data from kidney transplant recipients not included in the model development.

\section{Related work}
The standard method for measuring systemic exposure of drugs like tacrolimus is to obtain 8-12 blood samples spread over a dosing interval. This is challenging to implement in clinical practice, expensive, and not patient friendly. Hence, several methods requiring less frequent blood sampling have been developed for prediction of tacrolimus exposure. The majority of the methods aims to estimate the \ac{PK} parameters of the drug, and use these values to calculate the drug exposure in individual patients, based on 1-4 blood samples obtained within the first few hours after the previous dose. Population \ac{PK} modeling is a commonly used approach that estimates both the \ac{PK} parameter values and their variability within the population by using a \textit{maximum a posteriori} Bayesian technique~\cite{brooks2016popPKreview}.

Population \ac{PK} models can be useful in a clinical setting for predicting drug responses in individuals given a specific dosing schedule~\cite{storset2015improved}. Both parametric and non-parametric modelling are applied in the field of \ac{TDM}, with different advantages and limitations. Non-parametric population \ac{PK} models avoid the normality assumption for the \ac{PK} parameters. This is advantageous when the population of interest includes differently distributed sub-populations~\cite{neely2012Pmetrics}. Non-parametric population \ac{PK} modeling has shown promising results for predicting tacrolimus trough concentrations in kidney transplanted patients~\cite{storset2015improved} and for limited sampling \ac{AUC} determination~\cite{gustavsen2020tacdrop}. Despite their popularity, such models are time consuming and computationally expensive to train, and performance depends on how well the model structure captures \ac{PK} properties of the drug in the population of interest.

\Ac{ML} models are trained from data without being explicitly programmed. Model development hence does not require explicit assumptions regarding the \ac{PK} properties of a drug. In order to predict the \ac{AUC} of tacrolimus using \ac{ML}, \cite{niel2018artificial} trained an \ac{ANN} with one hidden layer on concentrations measured three hours after drug administration. Their dataset included $53$ observations, and the resulting \ac{ANN} successfully predicted the \ac{AUC} values, with superior performance compared to existing Bayesian models. Woillard et al.~\cite{woillard2021tacrolimus} used XGBoost to predict tacrolimus exposure in kidney transplant recipients for tacrolimus administered twice a day. Their models were trained and tested on data from $3,748$ and $1,249$ patients, respectively, and all showed better predictive power than existing population \ac{PK} models. However, neither~\cite{niel2018artificial} or~\cite{woillard2021tacrolimus} report prospective testing of their models. 

Existing studies applying \ac{ML} models for prediction of tacrolimus exposure indicate that \ac{ML} constitutes a powerful alternative to traditional approaches. In this study, we therefore develop \ac{ML} models that estimate tacrolimus exposure based on individual patient characteristics and measured drug concentrations.

\section{Data description}\label{sec:data}
The data used for model development includes tacrolimus measurements from $68$ individual adult kidney transplanted patients with $93$ unique visits at the clinic (some patients had visited the clinic more than once). The patients in the dataset had registered $8$ to $15$ drug concentrations for each dose interval. All patients administered tacrolimus twice a day, i.e., every $12$ hours. The data is obtained from four clinical studies performed at Oslo University Hospital - Rikshospitalet, Norway in the period 2011-2018~\cite{midtvedt2011advagraf,robertsen2015genTac,gustavsen2020dayNight,gustavsen2020tacdrop}. All studies were performed in accordance with the Helsinki Declaration and Good Clinical Practice and were approved by the local ethic committee (REK). All patients provided written informed consent for using the data in analyses like the present one.  
Characteristics of the $68$ patients in the development dataset are summarised in~\cref{tab:demographics}. The distribution is relevant for the overall transplant population at our center, with, e.g., more males and an average age in the mid 50'ies. 
The variation in \ac{TXT} is large, ranging from $12$ days to more than $15$ years. The morning dose of tacrolimus ranges from $1$ to $8$ mg. 

\begin{table}
   \scriptsize
    \caption{\label{tab:demographics}Demographics of the development dataset.
    }
    \centering
   \begin{tabular}{p{0.22\textwidth}p{0.05\textwidth}
    p{0.05\textwidth}p{0.08\textwidth}}  
    \toprule 
         Feature & N/Mean & Std & Min/Max  \\
         \midrule
         Male : Female & $53$ : $15$ & NA & NA \\
         Age (years) & $55$ & $14$ & $21/79$ \\
         Body weight (kg) & $78.4$ & $15.6$ & $47.8/116.0$\\
         Body mass index (kg/$m^{2}$) & $25.2$ & $3.9$ & $18.0/35.1$ \\
         Fat-free mass* (kg) & $60.4$ & $11.8$ & $35.3/79.1$ \\
         Hematocrit (\%) & $36.5$ & $3.6$ & $25.0/44.0$ \\
         \ac{TXT} (days) & $455$ & $1,004$ & $12/5,793$ \\
         Morning dose of tacrolimus (mg) & $3.0$ & $1.6$ & $1.0/8.0$ \\
         \bottomrule
         \multicolumn{4}{p{0.92\linewidth}}{Abbreviations: N = number of subjects; Std = standard deviation, NA = not applicable. *: Calculated using the Janmahasatian equation~\cite{janmahasatian2005FFMJ}. 
         } \\
    \end{tabular}
\end{table}

The data used for external testing of the final models includes tacrolimus measurements from seven adult kidney transplanted patients, all of them providing two dosing events. All patients had $12$ registered concentrations for each dose interval. The data was obtained from an ongoing clinical study performed at Oslo University Hospital - Rikshospitalet. The study was approved by the local ethical committee (reference number 146884) and the Norwegian Medicine Agency (EudraCT number 2020-002621-29). \Cref{tab:demographics_testset} shows the demographic characteristics for the seven patients. Again, most of the patients are male. The mean age is higher, and the \ac{TXT} is lower for the patients in the test set than for the patients in the development set. 

\begin{table}
    \scriptsize
    \caption{\label{tab:demographics_testset}Demographics of the external test dataset.
    }
    \centering
   \begin{tabular}{p{0.22\textwidth}p{0.05\textwidth}
    p{0.05\textwidth}p{0.07\textwidth}}  
    \toprule 
         Feature & N/Mean & Std & Min/Max  \\
         \midrule
         Male : Female & $5$ : $2$ & NA & NA \\
         Age (years) & $60$ & $16$ & $29/75$ \\
         Body weight (kg) & $81.8$ & $17.5$ & $54.8/101.9$\\
         Body mass index (kg/$m^{2}$) & $27.3$ & $4.6$ & $20.9/33.2$ \\
         Fat-free mass* (kg) & $57.1$ & $12.9$ & $36.6/71.6$ \\
         Hematocrit (\%) & $36.1$ & $4.2$ & $31.0/44.0$ \\
         \ac{TXT} (days) & $31$ & $6$ & $25/41$ \\
         Morning dose of tacrolimus (mg) & $2.9$ & $1.2$ & $1.0/4.5$ \\
         \bottomrule
         \multicolumn{4}{p{0.92\linewidth}}{Abbreviations: N = number of subjects; Std = standard deviation, NA = not applicable. *: Calculated using the Janmahasatian equation~\cite{janmahasatian2005FFMJ}. 
         } \\
    \end{tabular}
\end{table}

\section{Methodology}
\subsection{Data preprocessing}\label{sec:preprocessing}
Reference \ac{AUC} values are obtained using all measured blood concentrations in each individual, using the \texttt{makeAUC} function in the R-library Pmetrics~\cite{neely2012Pmetrics}. \ac{AUC} is calculated using the log-linear trapezoidal method for the measured tacrolimus blood concentrations within a dose interval. When a concentration at $12$ hours is missing, the trough concentration is used as an estimate of the concentration since all measurements were obtained in steady-state conditions.  

Development and test datasets are made using all available concentrations in the original datasets. The 16 time points with available concentrations are $0, 0.5, 1, 1.5, 2, 2.5, 3, 4, 5, 6, 7, 8, 9, 10, 11, 12$ hours after drug administration. For each time point, two features are created; one representing the drug concentration and one representing the time difference between the actual and ideal time points.
Missing concentrations are estimated using linear regression or log-linear regression: If missing concentrations come before the maximum concentration, linear regression using all available concentrations up to and including the maximum concentration, except the trough concentration, is used. If missing concentrations come after the concentration following the maximum concentration, log-linear regression using all available concentrations after the maximum concentration is applied. Linear regression is used for missing concentrations between the maximum and next available concentration.
\footnote{One event in the development dataset has only one measured concentration after the maximum concentration. For this event, the trough concentration is used as an estimate of the $12$-hour concentration before log-linear estimation of the missing concentrations is used to avoid constant concentration over time.} 

Additional patient characteristics included in the data are body weight and height, body mass index, body surface area, sex, hematocrit (percentage of red blood cells), \ac{TXT}, type of assay used in the blood sample bioanalyses, and an error term representing the uncertainty of the bioanalyses. In total, each data point is characterised by 43 features.

\subsection{Experiments}
Various feature combinations were explored for predicting \ac{AUC}. To increase applicability in the clinic, we focus on trough concentrations and concentrations measured one and three hours after drug administration. 

Most \ac{ML} models require all features used during development to be present when making predictions. However, the patient might not be available for blood sampling at exactly the desired time. We therefore relax the time constraint of the third concentration by adding additional features, representing the time deviation from three hours after dose administration. 
Five different time points are considered for measuring the last concentration during model development: 2, 2.5, 3, 4 and 5 hours after drug administration. Consequently, events for all these scenarios are created for each of the 68 patients, resulting in 340 events in total.
Two different approaches are tested in order to create \ac{ML} models that handle deviations in time for measuring the third concentration. The first approach provides the measured concentration and the time deviation away from three hours directly to the model. The second approach estimates the concentration at three hours based on the other available drug concentrations. The estimated concentration is then provided to the \ac{ML} model together with a binary feature flagging whether the concentration was measured on time or was estimated. 

For the first approach, the following combinations of features are explored: 
\begin{itemize}
  \item Feature representing time difference between three hours and actual time
  \item Features for exact times for all three concentrations
  \item Combination of the two first points and features for body weight and height, sex, body mass index, body surface area, hematocrit and type of assay for the analysis
  \item Combination of the two first points without adding more features
\end{itemize}

For the second approach, the following methods are explored for estimating the concentration at three hours:
\begin{itemize}
    \item Linear regression between the concentration measured at one hour and the latest concentration
    \item Linear regression, but the one-hour concentration and last concentration are swapped if the last concentration is higher than the  one-hour concentration. This is done to prevent the estimated concentration at three hours to become unrealistically high, as the maximum concentration most likely arise before three hours. The approach is referred to as `reverse linear regression' in the following sections
    \item Log-linear regression between the latest concentration and the trough concentration, representing the 12-hour concentration
    \item Estimations from an earlier developed population \ac{PK} model~\cite{aasberg2013popPKmodel} receiving patient characteristics, the three drug concentrations and corresponding time points
\end{itemize}

Leave-one-patient-out \ac{CV} is performed for all the \ac{ML} experiments. When the dataset is small and no external test set is available, leave-one-out \ac{CV} is regarded as best practice. Only the first visit for each patient is applied, resulting in $68$ events for fixed time points, and $340$ events for the flexible time points.
Before prospective testing, the \ac{ML} models are trained on all available events in the development dataset. If a patient has been investigated several times, all events are included, treating each event as an independent patient. This results in $93$ and $465$ events for the models trained on fixed and flexible time points, respectively. The models are then tested on the external test set, which was not applied for model development.

The XGBoostRegressor algorithm~\cite{Chen2016xgboost} is used to build the \ac{ML} models. For experiments using leave-one-patient-out, the best hyperparameter combinations identified in preliminary experiments are used for training the \ac{ML} models. For the models used in prospective testing, hyperparameter search using GroupKFold \ac{CV} from sklearn~\cite{scikit-learn} is performed to find the best combination of hyperparameter values. GroupKFold ensures that events from the same patient are always placed together in either the training or validation set during \ac{CV}.  

The performance of the \ac{ML} models are compared to an established population \ac{PK} model~\cite{aasberg2013popPKmodel}. The population \ac{PK} model receives the same concentrations as the \ac{ML} models and the corresponding exact sampling time points. The population \ac{PK} model automatically handles deviations in time. For flexible time points, the model therefore receives drug concentrations and corresponding deviating times instead of using the approaches outlined above. Moreover, the population \ac{PK} model receives the dose of the drug, body weight and height, \ac{TXT}, hematocrit, sex, body mass index and fat-free mass. These features must be present for the model to run the analysis. In addition, the DummyRegressor from sklearn~\cite{scikit-learn} is applied as a baseline model. The DummyRegressor always predicts the mean target value of the training dataset.

\subsection{Model evaluation}
Relative \ac{RMSE} is used for model evaluation and model selection. Relative \ac{RMSE} is calculated by dividing the \ac{RMSE} by the mean target value of the validation or test data for leave-one-patient-out \ac{CV} and prospective testing, respectively. The percentages of patients with absolute \acp{PE} above 15\% and 10\% of the target value are also calculated. For a model to be clinically applicable, the number of observations with absolute \acp{PE} above $15 \%$ should not exceed $15 \%$. As a rule of thumb, models with relative \acp{RMSE} below $10 \%$ are generally regarded as `good', while values below $15 \%$ are regarded as `okay'~\cite{gustavsen2020tacdrop}. \ac{SHAP} is a popular explanation method that approximates Shapley values~\cite{lundberg2019SHAP}. Shapley values originate from game theory, and the value for a given player, or feature, reflects its contribution to the total payoff, or prediction~\cite{Shapley1953}. We use \ac{SHAP} during prospective testing to identify the most important features for our \ac{ML} models. If the important features are clinically plausible, this can strengthen the trust in the models.

\section{Results}
\subsection{Leave-one-patient-out cross-validation}
The results from the leave-one-patient-out \ac{CV} experiments on the development dataset are shown in~\cref{tab:loo_1}. 
All XGBoost models perform better than the DummyRegressor baseline model. The best model performance is achieved when all available features are included. When the last time point is not fixed at three hours after drug administration, model performances drop, meaning that the \ac{ML} models are less robust to deviations in sampling time. For flexible time points, the lowest relative \ac{RMSE} is achieved when the time difference from three hours is combined with features describing the time point in minutes after the previous dose, as well as other patient characteristics.

\begin{table}
    \scriptsize
    \caption{\label{tab:loo_1}ML model results with leave-one-out CV.
    }
    \centering
   \begin{tabular}{p{0.44\columnwidth}p{0.12\columnwidth}p{0.11\columnwidth}p{0.11\columnwidth}} 
    \toprule 
        Type of model & \% relative RMSE & \% outside $10\%$ PE & \% outside $15\%$ PE\\
        \midrule
        Fixed time points \\
        \midrule
        $16$ concentrations & $9.45$ & $13.24$ & $5.88$\\
        3 concentrations & $10.04$ & $19.12$ & $7.35$\\
        3 concentrations, less features & $11.15$ & $23.53$ & $8.82$\\
        \midrule
        Flexible time points \\
        \midrule
        \textbf{First approach} \\
         Feature for time difference & $16.88$ & $37.65$ & $17.94$ \\ 
         Features for exact time & $17.36$ & $42.94$ & $24.41$\\ 
         Combination of two first & $16.65$ & $35.00$ & $14.71$\\ 
         Combination of two first, more features & $15.98$ & $44.12$ & $21.18$ \\
         \midrule
         \textbf{Second approach} \\
         Linear regression & $17.20$ & $36.76$ & $19.71$ \\ 
         `Reverse' linear regression & $18.16$ & $50.29$ & $31.47$ \\
         Log-linear regression & $16.25$ & $34.12$ & $17.06$\\
         Population PK model concentration & $16.29$ & $37.35$ & $19.71$\\
         \midrule
        Baseline model & $29.23$ & $69.12$ & $42.65$ \\
        \bottomrule
    \end{tabular}
\end{table}

\subsection{Prospective testing}
\Cref{tab:test_resultsFixed} shows the results from prospective  external testing of the \ac{ML} models. As for the development data, all models outperform the baseline model, and the best results are obtained when all features are included. For the models that are trained on three concentrations, performances slightly drop when more features are included, as opposed to the results from internal model evaluation. The best approach for flexible time points is when missing concentrations at three hours post-dose are estimated by the population \ac{PK} model. Based only on the criteria for clinical applicability, four of the \ac{ML} models presented here could be used in the clinic, from which one handles deviations in the last time point. 

\begin{table}
    \scriptsize
    \caption{\label{tab:test_resultsFixed}ML model results on external test data. 
    }
    \centering
   \begin{tabular}{p{0.44\columnwidth}p{0.12\columnwidth}p{0.11\columnwidth}p{0.11\columnwidth}} 
    \toprule 
        Type of model & \% relative RMSE & \% outside $10\%$ PE & \% outside $15\%$ PE\\
        \midrule
        Fixed time points \\
        \midrule
        $16$ concentrations & $3.99$ & $0.0$ & $0.0$ \\
        3 concentrations & $9.85$ & $28.57$ & $7.14$  \\
        3 concentrations, less features & $9.72$ & $28.57$ & $7.14$ \\
        \midrule
        Flexible time points \\
        \midrule
        \textbf{First approach} \\
        Feature for time difference & $13.17$ & $45.71$ & $21.43$ \\ 
        Features for exact time & $14.79$ & $47.14$ & $32.86$ \\ 
        Combination of two first & $13.29$ & $40.00$ & $27.14$ \\ 
        Combination of two first, more features & $14.66$ & $47.14$ & $24.29$ \\
        \midrule
        \textbf{Second approach} \\
        Linear regression & $14.70$ & $52.86$ & $27.14$  \\ 
        `Reverse' linear regression & $14.90$ & $55.71$ & $31.43$  \\
        Log-linear regression & $11.74$ & $42.86$ & $17.14$ \\
        Population PK model concentration & $11.19$ & $40.00$ & $11.43$ \\
        \midrule
        Baseline model & $19.75$ & $71.43$ & $50.00$ \\
        \bottomrule
    \end{tabular}
\end{table}

\begin{figure*}
    \centering
    \includegraphics[trim=0cm 6.82cm 0cm 6.9cm, clip,width=\linewidth]{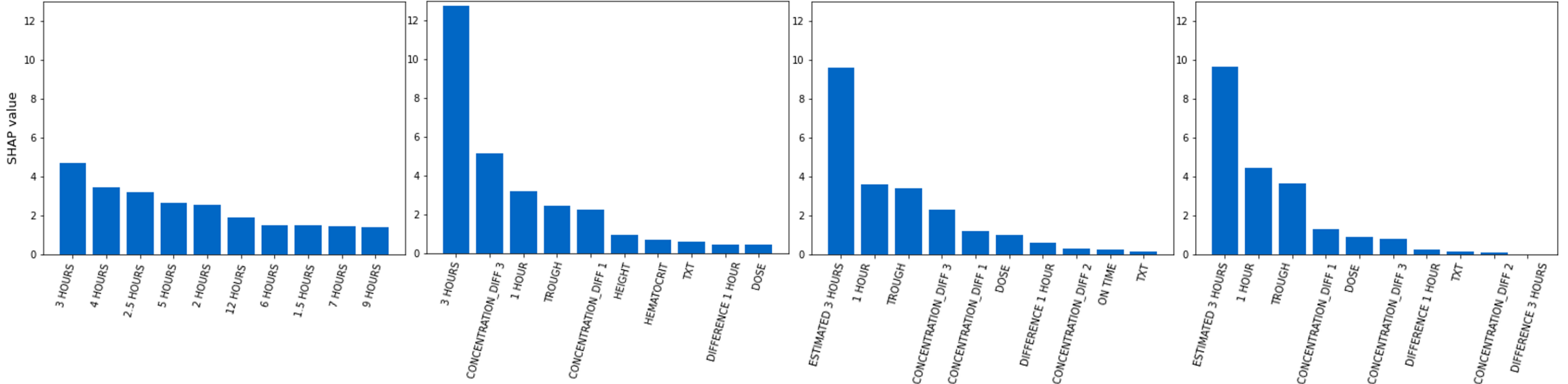}
    \caption{From left to right: \ac{SHAP} values for the ML models trained on all concentrations, on three concentrations and additional features, flexible times where missing concentrations at three hours are estimated by the population PK model and flexible times where concentrations missing at three hours are imputed using log-linear regression, respectively. Only the top ten features are included for convenience.}
    \label{fig:SHAP_plots}
\end{figure*}

The ranking of \ac{SHAP} values is investigated for the \ac{ML} models when making predictions on the test set. For all the models, the top-ranked features represent drug concentrations. The best performing \ac{ML} model, which received $16$ drug concentrations, treats the three-hour concentration as the most important feature. The first feature not representing either a drug concentration or a part of the error term for the analysis is ranked as number $17$ (hematocrit). Since the \ac{AUC} calculations are based only on measured drug concentrations, it makes sense that these are regarded as most important by the model. In fact, it might be that the best performing model more or less learns the trapezoidal method. \ac{SHAP} values for the best performing model are shown in~\cref{fig:SHAP_plots}. For the two models trained on three drug concentrations at fixed times, the three-hour concentration is ranked as either number one or two, respectively, competing with the feature measuring the concentration difference between three and zero hours. Trough and one-hour concentrations are also highly ranked. Height, hematocrit and \ac{TXT} are the most important features not representing drug concentrations. \ac{SHAP} values for the model trained on three drug concentrations and additional features are shown in~\cref{fig:SHAP_plots}.
Similar results are achieved for the models developed with a flexible time point for the last concentration. The model where missing three-hour concentrations are imputed by the population \ac{PK} model, treats the three-hour concentration as the most important feature, followed by the one-hour and trough concentrations. The same is true for the model where missing three-hour concentrations are estimated using log-linear regression. \Cref{fig:SHAP_plots} plots the \ac{SHAP} values for both models. 

\subsection{Population pharmacokinetic model}
The results from the \ac{AUC} estimations on the development data using the previously developed population \ac{PK} model are included in the upper part of~\cref{tab:pop_model}. As for the \ac{ML} models, the best performance is achieved when the number of provided concentrations is the largest. The performance drops when only three concentrations are available. The model performance is not significantly different between predictions made on fixed and flexible time points, which is in contrast to the \ac{ML} models. 
Corresponding results on the test data are shown in the lower part of~\cref{tab:pop_model}. The trend is similar to the trend on the development data. However, the relative \ac{RMSE} and percentages of samples with \acp{PE} outside $10$ and $15 \%$ are significantly lower when the population \ac{PK} model receives $16$ concentrations. Overall, the population \ac{PK} model performs better than the \ac{ML} models. 

\begin{table}
    \scriptsize
    \caption{\label{tab:pop_model}Population PK model results on development and test data. 
    }
    \centering
   \begin{tabular}{p{0.29\columnwidth}p{0.13\columnwidth}p{0.13\columnwidth}p{0.13\columnwidth}} 
    \toprule 
        & \% relative RMSE & \% outside $10\%$ PE & \% outside $15\%$ PE \\
        \midrule
        Development data \\
        \midrule
        16 concentrations & $6.92$ & $13.24$ & $5.88$ \\
        3 fixed time points & $9.21$ & $26.47$ & $10.29$  \\
        3 flexible time points & $9.09$ & $26.47$ & $10.29$ \\
        \midrule
        Test data \\
        \midrule
        16 concentrations & $1.88$ & $0.00$ & $0.00$ \\
        3 fixed time points & $9.05$ & $28.57$ & $14.29$ \\
        3 flexible time points & $8.67$ & $28.57$ & $14.29$ \\
        \bottomrule
    \end{tabular}
\end{table}

\section{Discussion}
Our results show that the population \ac{PK} model tends to outperform the \ac{ML} models, with similar performance for the population \ac{PK} and \ac{ML} models for fixed concentrations.  
This can have several reasons. Because the population \ac{PK} model learns the underlying \ac{PK} processes leading to the observed drug concentrations that are used to estimate the \ac{AUC} values, it might capture relationships not learned by the \ac{ML} models. Moreover, the population \ac{PK} model handles exact sample times and estimates missing concentrations without further manipulation. This could be advantageous, especially when the last concentration sampling time deviates from three hours. This is reflected in the results, as the population \ac{PK} model clearly outperforms the \ac{ML} models for flexible time for the three-hour concentration.

According to the external test results, the best performing \ac{ML} model for flexible times estimates the concentration at three hours using the population \ac{PK} model. The population \ac{PK} model performs well when the last concentration is not measured at exactly three hours post-dose, which might explain its good performance. A drawback is that the workflow is more complex than for the other approaches, as the population \ac{PK} model must pass the estimated values to the \ac{ML} model before making predictions. The second best performing approach, using log-linear regression to estimate the three-hour concentration, is faster and less computationally expensive. However, the percentage of samples with \acp{PE} outside $15 \%$ of the reference \ac{AUC} values is above the recommended value of $15 \%$.

According to \ac{SHAP}, features representing drug concentrations and differences between drug concentrations at different times are most important for all our \ac{ML} models. This makes sense, as the \ac{AUC} calculations are based on drug concentrations. Moreover, high concentrations and large differences between the three-hour concentration and trough concentration lead to higher predicted \ac{AUC}. This is also reasonable considering the tacrolimus blood concentration profile, as higher concentrations and large increases in concentrations leads to higher \ac{AUC}. 
Drug dose is not among the top features. This supports the fact that different patients might experience different drug exposures although they administer the same dose of tacrolimus. Consequently, the tacrolimus dose is not a good predictor of the \ac{AUC}, which is why obtaining a good prediction model for individualizing dosing is important. Apart from features related to drug concentrations, hematocrit is ranked relatively high for some of the best performing models. Higher hematocrit contributes to higher predicted \ac{AUC}, although the effect is not large according to \ac{SHAP}. Tacrolimus binds extensively to red blood cells~\cite{venkataramanan1995tacRedBloodCells}, and it is only the free concentration that is available for elimination. Consequently, higher hematocrit results in less free tacrolimus, lower total clearance and higher total \ac{AUC}. 
Prior research suggests that \ac{TXT} affects the drug exposure of tacrolimus~\cite{staatz2004tacPKreview}. However, \ac{TXT} is not an important feature for the \ac{ML} models according to \ac{SHAP}. All patients in the test set are transplanted less than two months ago, and the variation in \ac{TXT} is relatively small. Consequently, the effect of \ac{TXT} on tacrolimus \ac{AUC} might not be detectable in the current test set.  

This study has some limitations.
First, the log-linear trapezoidal method is used to calculate the reference \ac{AUC} values. This is a conservative recognized method that does not anticipate any information about the population or data distributions. It provides accurate \acp{AUC} for patients with frequently sampled data, but is less accurate when the number of available concentrations per individual is limited. For patients without registered concentrations at $12$ hours, the trough concentration is used as an estimate. This introduces some uncertainty, making the reference \ac{AUC} values slightly biased. Still, our data includes at least $8$ drug concentrations for each event, which should yield reasonably accurate \ac{AUC} estimations. 

For log-linear estimation of missing concentrations, we use all available concentrations in the relevant time interval. Using only the two closest concentrations might give more accurate estimations, considering the \ac{PK} properties of tacrolimus, and should be tested as future work.

When creating the datasets applied for flexible time, 2, 2.5, 3, 4 and 5 hours are considered. In a real-life setting, concentrations are not necessarily measured at these times. However, it would not be feasible to include all possible time points during model development. We believe that these five times are representative and cover most time intervals likely to be used in clinical practice. 
Moreover, we only test flexible times for the last concentration. In future work, deviations for trough and one-hour concentrations should also be explored. 

With population \ac{PK} models, it is possible to investigate the underlying \ac{PK} properties of drugs. They are more interpretable than \ac{ML} models, but also tedious to develop, not necessarily supporting automatic hyperparameter tuning, and requiring much computational power. 
Our \ac{ML} models are faster to train and run and support automatic hyperparameter optimization. They are less computationally heavy, but still achieve results comparable to the population \ac{PK} model. They are, however, not suitable for describing the \ac{PK} properties of drugs. It is, however, still possible to gain insight into the model using explainable artificial intelligence, as demonstrated here using \ac{SHAP}. Such analyses can increase trust in the \ac{ML} models, as our findings suggest that our models are able to reflect some of the processes that affect the \ac{PK} of tacrolimus. 

\section{Conclusion}
To conclude, we have developed \ac{ML} models that predict tacrolimus exposure in kidney transplanted patients. Results from prospective testing are promising and indicate that \ac{ML} models can have predictive performances at the same level as more established population \ac{PK} models. For future work, we plan to increase the dataset size using synthetic data generation and test if this can improve performance or even compensate for missing values.

\bibliographystyle{IEEEtran} 
\small
\bibliography{bibliography}
\end{document}